# GLCM-Based Feature Combination for Extraction Model Optimization in Object Detection Using Machine Learning

Florentina Tatrin Kurniati[1,2], Irwan Sembiring[1], Adi Setiawan[3], Iwan Setyawan[4], Roy Rudolf Huizen[5]

[1]Faculty of Information Technology Universitas Kristen Satya Wacana,
Jl. Diponegoro No.52-60, Salatiga, Jawa Tengah 50711
[2]Faculty of Informatics and Computer, Institut Teknologi dan Bisnis STIKOM Bali, Indonesia
Jl. Raya Puputan No.86, Denpasar Bali 80234
[3]Faculty of Science and Mathematics, Universitas Kristen Satya Wacana
Jl. Diponegoro No.52-60, Salatiga, Jawa Tengah 50711
[4]Faculty of Electronics and Computer Engineering Universitas Kristen Satya Wacana Salatiga
Jl. Diponegoro No.52-60, Salatiga, Jawa Tengah 50711
[5]Department of Magister Information System, Institut Teknologi dan Bisnis STIKOM Bali, Indonesia
Jl. Raya Puputan No.86, Denpasar Bali 80234

## ARTICLE INFO

## ABSTRACT

In the era of modern technology, object detection using the Gray Level Co-occurrence Matrix (GLCM) extraction method plays a crucial role in object recognition processes. It finds applications in real-time scenarios such as security surveillance and autonomous vehicle navigation, among others. Computational efficiency becomes a critical factor in achieving real-time object detection. High computing time delays can cause overall system failure. Hence, there is a need for a detection model with low complexity and satisfactory accuracy. This research aims to enhance computational efficiency by selecting appropriate features within the GLCM framework. Two classification models, namely K-Nearest Neighbours (K-NN) and Support Vector Machine (SVM), were employed, with the results indicating that K-NN outperforms SVM in terms of computational complexity. Specifically, K-NN, when utilizing a combination of Correlation, Energy, and Homogeneity features, achieves a 100% accuracy rate with low complexity. Moreover, when using a combination of Energy and Homogeneity features, K-NN attains an almost perfect accuracy level of 99.9889%, while maintaining low complexity. On the other hand, despite SVM achieving 100% accuracy in certain feature combinations, its high or very high complexity can pose challenges, particularly in real-time applications. Research contribution to improving computational efficiency in object detection using the GLCM method and KNN and SVM classification models to achieve high accuracy with low complexity. Therefore, based on the trade-off between accuracy and complexity, the K-NN model with a combination of Correlation, Energy, and Homogeneity features emerges as a more suitable choice for real-time applications that demand high accuracy and low complexity. This research provides valuable insights for optimizing object detection in various applications requiring both high accuracy and rapid responsiveness.

**Corresponding Author**:

Florentina Tatrin Kurniati, Faculty of Information Technology Universitas Kristen Satya Wacana
Jl.Diponegoro No.52-60, Salatiga, Jawa Tengah 50711
Email: 982022026@student.uksw.edu

## 1. INTRODUCTION

In today's modern technological era, object detection systems have been implemented in various ways including security surveillance, autonomous car navigation, facial detection systems, and industrial automation [1], [2] The challenge of the system in a real-time environment is time delay problems caused by algorithm complexity [3], [4]. Object detection methods include the GLCM (Gray-Level Co-occurrence Matrix) method used in pixel analysis to improve accuracy. This method utilizes attributes such as energy, contrast, entropy, variance, correlation, and homogeneity in the object extraction. Using an iterative counting approach, GLCM considers certain pairs of pixels with defined intensities. In addition, this method also considers special angles such as 0°, 45°, 90°, and 135° in pixel analysis [5], [6]. The use of various orientation angles in the GLCM method in pixel analysis allows for more detailed information about the structure and characteristics of the object being analyzed [7], [8]. In real-time systems, computing problems are often the main challenge. Long computing processes can cause latency, namely delays in sending or receiving information, which has the potential to reduce the effectiveness of object detection systems [9]. Handling computational problems is the focus of optimization in increasing the efficiency and effectiveness of this object detection system in a real-time environment [10].

The object detection process consists of recognizing and grouping objects based on features. The extraction process is the stage of taking information from features for detection. Features as input to machine learning algorithms used to train models to recognize and identify objects. Gray Level Co-occurrence Matrix (GLCM) is an effective method for feature extraction. GLCM produces various types of features such as contrast, correlation, energy, homogeneity, and entropy [11]–[14]. Each of these features plays an important role in the object detection process. For example, contrast features include variations in color and intensity between one pixel and its neighbors, while energy features measure the degree of uniformity or homogeneity of an object and so on. The GLCM feature extraction method in machine learning requires a long computation time [15]–[18]. This is influenced by many calculations and analyses that can make the model less than optimal. Computational optimization is important to ensure that the system can work effectively and efficiently, while the machine learning model can train itself properly and produce accurate and reliable results. Based on this, this study focuses on increasing computational time with the feature selection approach in GLCM for the object detection process. Research contribution to improving computational efficiency in object detection using the GLCM extraction method with feature variants. The next contribution combines KNN and SVM classification models to obtain high accuracy but with low complexity.

## 2. METHODS

This test uses a dataset totaling 100,000 with a division of 90% for training and 10% for testing. The data used is image data of triangles, squares and circles that are rotated or change dimensions. Using these data, detection with high accuracy and low complexity is the proposed model. The developed test model combines the features in GLCM.

The goal is to get a combination of features to reduce the complexity of the algorithm. Each pair of features is calculated computation time and accuracy value. To combine as many as 5 features are used, with each combination of 2 and 3, a total of 20 feature combination variants. Each feature combination variant is analyzed at angles of 0°, 45°, 90°, and 135°. The angle selector is to obtain features from all orientations. In this process, the classification method used is K-NN and SVM, the GLCM-Based Feature Combination model is shown in Fig. 1. Features with 2 and 3 combinations and KNN or SVM classification, aims to determine the effect of combination pairs and classification in machine learning on accuracy and complexity. The stages of the object detection process are pre-processing for resizing, and extraction with a combination of GLCM features (2 combinations and 3 combinations) with the extraction results classified by K-NN and SVM. The results of the model are accuracy and complexity.

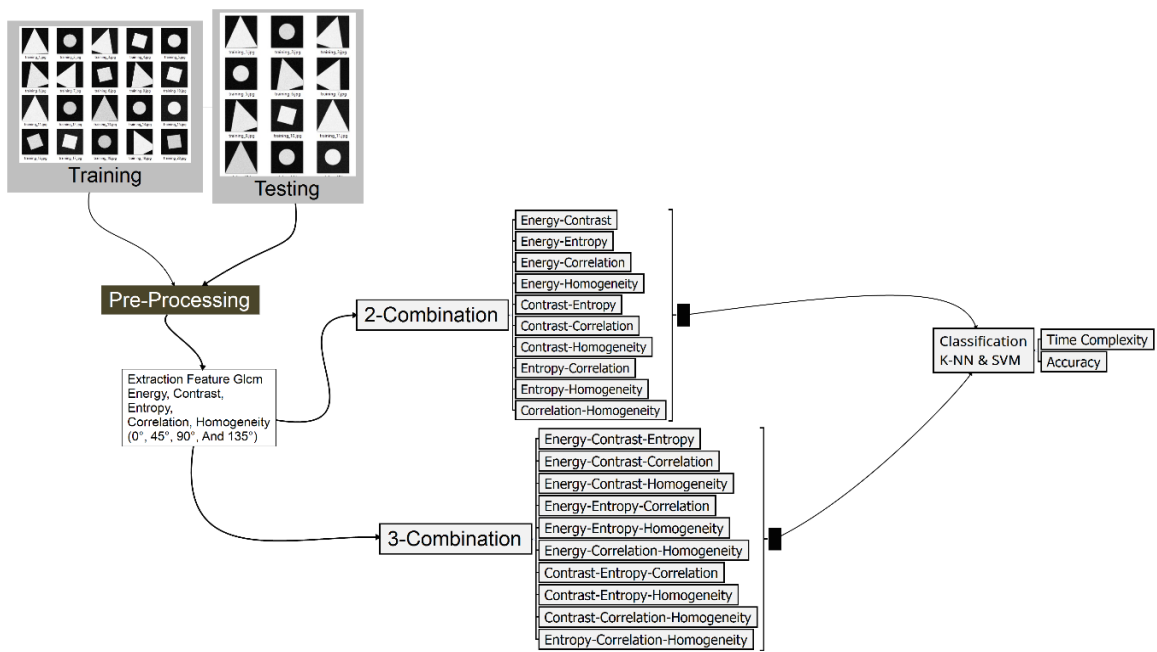

**Fig. 1.** GLCM-Based Feature Combination Flow Diagram

## 2.1. Algorithm Complexity

Gray Level Co-occurrence Matrix (GLCM) is a statistical method that measures image texture by considering the spatial relationship between pixels, used in texture analysis and pattern recognition. The complexity of the GLCM algorithm is related to the number of operations required to produce an $O(n^2)$ matrix, where n is the number of pixels in the image [19]. This notation is the upper limit of the time needed to run the algorithm with input size n. Complexity is evaluated for the worst, average or best cases, as shown in Table 1 Complexity Algorithm [20].

**Table 1.** Notasi big O Complexity Algorithm

| No. | Feature | Complexity (big-O) |
|---|---|---|
| 1. | Energy | $O(n^2)$ |
| 2. | Contrast | $O(n^2)$ |
| 3. | Homogeneity | $O(n^2)$ |
| 4. | Entropy | $O(n^2)$ |
| 5. | Correlation | $O(n^2)$ |

The GLCM features energy, contrast, homogeneity, entropy and correlation have a (big-O) complexity notation of $O(n^2)$. In $O(n^2)$ notation it refers to the time complexity required to construct the gray-level co-occurrence matrix of an image. for the notation, n represents the number of pixels in the image. image scanning and calculation of the frequency of pixel pairs with a certain intensity at a certain distance and orientation, generally two loops (nested loops), one for rows and one for image columns, so the complexity is $O(n^2)$ [21]–[24].Each iteration, the algorithm will calculate the relationship between pixels and their neighbors. So that the image size increases, the time needed to calculate the GLCM matrix will increase quadratically. As an illustration for a size of 100×100 pixels (10,000 pixels), it is necessary to process more than 100 million (10,000^2) operations to calculate the GLCM matrix [25]–[27].

## 2.2. Feature Combination GLCM

The model in Figure 1 is an image texture analysis method by extracting features. The extraction method uses a combination of GLCM features (2 combinations and 3 combinations). A combination of feature pairs will be trained with the KNN and SVM classification models. Processing uses machine learning, a combination that gives the best results with the highest classification accuracy and low computation time. The combination of GLCM features used, namely energy, is a measure that describes the uniformity of elements in the GLCM.

In (1) the energy value will be high if the elements in the matrix tend to have the same or uniform values. Energy can be calculated using the following formula [28], [29]:

$$Energy = \sum_i \sum_j P(i,j)^2 \qquad (1)$$

Equation (2) contrast, also known as inertia, is a measure that describes the degree to which the intensity of different pixels in an image varies. If the contrast is high, it means that there is a lot of variation in pixel intensity; conversely, low contrast indicates less variation. Contrast can be calculated by the following formula [30], [31];

$$Contrast = \sum_i \sum_j (i-j)^2 * P(i,j) \qquad (2)$$

Equation (3) homogeneity is a measure that reflects the degree to which the matrix elements are close to the diagonal of the GLCM matrix. The homogeneity value will be high if the matrix elements that have high values are located near the diagonal. Homogeneity can be calculated using the following formula [29], [32]:

$$Homogeneity = \sum_i \sum_j \frac{P(i,j)}{1+(i-j)^2} \qquad (3)$$

Equation (4) entropy is a measure of the complexity of the texture or information contained in an image. Higher entropy indicates higher texture or information complexity. Entropy can be calculated using the following formula [30], [33];

$$Entropy = -\sum_i \sum_j P(i,j) * log(P(i,j)) \qquad (4)$$

Equation (5) correlation is a measure that describes the extent of the relationship between two pixels. A high correlation indicates a strong dependency between two pixels. Correlation can be calculated using the following formula [34], [35];

$$Correlation = \frac{\sum_i \sum_j [ij * P(i,j)] - \mu_x * \mu_y}{\sigma_x * \sigma_y} \qquad (5)$$

where $i$ and $j$ is co-occurrence matrix indices, $P(i,j)$ is the element of the co-occurrence matrix at position $(i,j)$, $\mu x$ and $\mu y$ is the average row and column weights of the co-occurrence matrix, $\sigma x$ and $\sigma y$ is the standard deviation of the row and column weights of the co-occurrence matrix.

These features are combined 2 and 3, the combination results are shown in Table 2. Each feature combination is used to extract the image. This combination produces a total of 20 feature variants.

Table 2. GLCM Feature Combination

| No | 2 Combination | 3 Combination |
|---|---|---|
| 1 | Energy, Contrast | Energy, Contrast, Homogeneity |
| 2 | Energy, Homogeneity | Energy, Contrast, Entropy |
| 3 | Energy, Entropy | Energy, Contrast, Correlation |
| 4 | Energy, Correlation | Energy, Homogeneity, Entropy |
| 5 | Contrast, Homogeneity | Energy, Homogeneity, Correlation |
| 6 | Contrast, Entropy | Energy, Entropy, Correlation |
| 7 | Contrast, Correlation | Contrast, Homogeneity, Entropy |
| 8 | Homogeneity, Entropy | Contrast, Homogeneity, Correlation |
| 9 | Homogeneity, Correlation | Contrast, Entropy, Correlation |
| 10 | Entropy, Correlation | Homogeneity, Entropy, Correlation |

### 2.3. Classification of K-NN and SVM

The K-Nearest Neighbors (K-NN) and Support Vector Machines (SVM) methods were used in this study to classify the training dataset and test data. Both classification methods have been frequently used in machine learning and have proven to be reliable. For the K-NN method, objects tend to have the same class as their neighbors [36], [37]. The K-NN algorithm identifies the closest 'K' observations from the training data and classifies these objects to the class that occurs most often from the closest 'K' samples. This method is intuitive and easy to implement but can be slow if the size of the training data is very large [34]. While the SVM method looks for the optimal hyperplane separating the data into two classes [38], [39]. Hyperplane as a subspace with one dimension less than the original space [40], [41]. The best hyperplane is selected from each maximum class. If the data cannot be separated linearly, SVM uses a kernel trick to map the data to a higher dimension. For the K-NN algorithm shown in Table 3. As for the SVM algorithm, shown in Table 4. Selection of the classification algorithm and the right feature combination can contribute to increasing the value of accuracy and optimal performance in image classification.

**Table 3.** K-Nearest Neighbors Algorithm

| | |
|---|---|
| Step 1 | Determine the K parameter as the number of nearest neighbors to be used in the classification or regression process. |
| Step 2 | Determine the distance between the data to be classified and each existing training data using the distance metric<br>Euclidean Distance:<br>D(x, y) = sqrt((x1 - y1)^2 + (x2 - y2)^2 + ... + (xn - yn)^2) |
| Step 3 | Identify the nearest K neighbors based on the distance that has been calculated. |
| Step 4 | Calculates the frequency of each class (for classification) or the average value (for regression) of the nearest neighbors. |
| Step 5 | Menetapkan data yang akan Assign data to be classified or predicted to the class with the highest frequency (classification) or the highest average value (regression). |

**Table 4.** Support Vector Machines Algorithm

| | |
|---|---|
| Step 1 | Choosing the appropriate kernel function (linear)<br>Linear Kernel: K(x, y) = x * y |
| Step 2 | Transforming the training data into a higher feature space (nonlinear transformation) using the selected kernel function. |
| Step 3 | Identify the best hyperplane that separates the two classes by the maximum margin limit. This hyperplane is usually known as the maximum margin hyperplane. |
| Step 4 | Define support vectors, i.e. training data that are at the margins or misclassified. |
| Step 5 | Using the identified hyperplane and support vectors for new data classification. |

## 3. RESULTS AND DISCUSSION

Test results with a dataset of 100,000 with a training composition of 90,000 and testing of 10,000. The first extraction test uses a combination of 2 features. In this test, the results are shown in Fig. 2. The combination of the two features shows the classification with SVM for features "contrast, correlation" and "contrast, energy" provides the highest accuracy, while "energy, homogeneity" gives the lowest accuracy. The contrast feature has a positive effect on accuracy, while the combination of energy and homogeneity has a negative effect. Classification using K-NN values high accuracy for almost all feature combinations, except "contrast, homogeneity" and "contrast, energy", where the accuracy is slightly lower. This shows that the contrast feature may hurt accuracy but not significant.

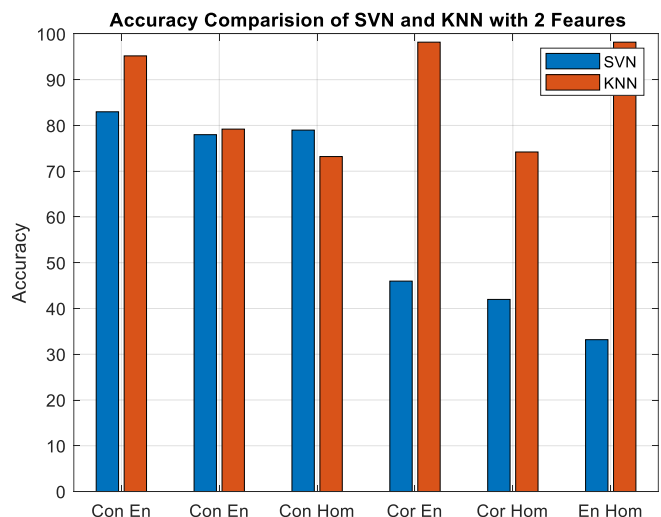

**Fig. 2.** Combination of two features with SVM and K-NN classification

The test results with the 3 combinations are shown n Fig. 3, for classification with SVM, the Entropy feature improves accuracy, especially in the combinations "contrast, correlation, entropy", "contrast, energy, entropy", and "contrast, homogeneity, entropy". As for the K-NN classification, all feature combinations are obtained. Therefore, it is difficult to determine which features have a negative or positive influence on accuracy.

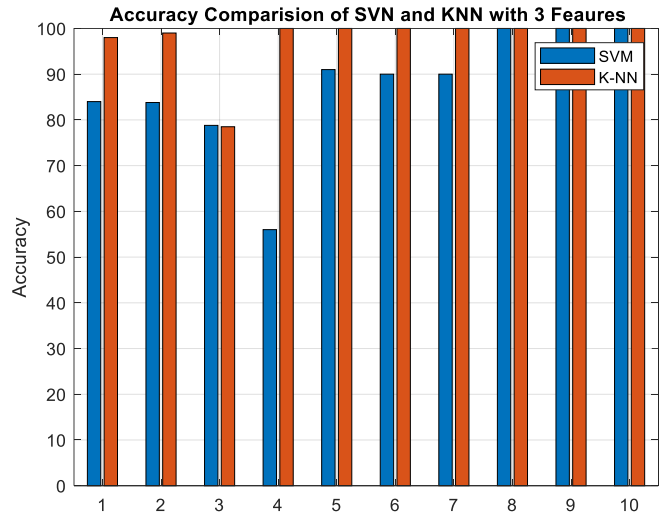

**Fig. 3.** Combination of three features with SVM and k-NN classification

Based on the test results, the average values for combinations of 2 features and 3 features are shown in Table 5. K-NN classification averaged 86.37% for 2 combinations and 96.24% for 3 combinations. As for SVM, the average of 2 combinations is 60.18% and the average of 3 combinations is 83.90%.

**Table 5.** Average accuracy data

| Feature Combination | SVM | K-NN |
|---|---|---|
| 2 | 60.18% | 86.37% |
| 3 | 83.90% | 96.24% |

The accuracy value of each combination compared to the complexity of the results for the 2 feature combinations is shown in Fig. 4. The energy feature complexity value has an increased contribution compared to other features as shown in Fig. 4 for variant 2 of the 1st to 5th feature combinations. SVM classification seems to require more computation time than K-NN. The difference between the two is very significant. The K-NN classification looks complicated for shorter object detection than SVM.

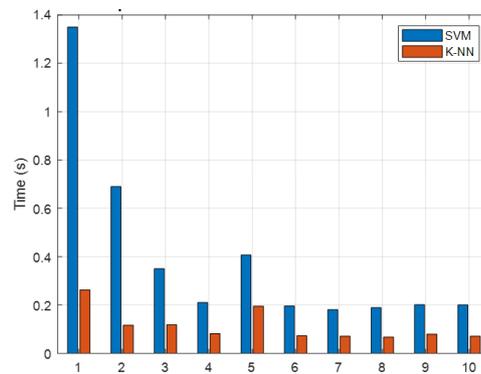

**Fig. 4.** Complexity combination of two features

Testing using 3 combinations is shown in Fig. 5. The SVM model exhibits a very high level of complexity. In the context of real-time systems, this is less than ideal because it can cause latency in system response. As for Energy and Homogeneity, the KNN classification has a low level of complexity compared to SVM. For the combination of 3 features, SVM shows a high or very high level of complexity, which may make it less suitable for use in real-time systems. In contrast, KNN consistently shows a moderate level of complexity.

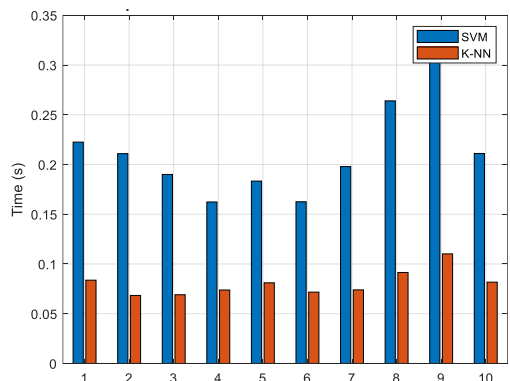

**Fig. 5.** Complexity combination of three features

This suggests that KNN may be more suitable for use in real-time systems. K-NN with Correlation, Energy, and Homogeneity features has 100% accuracy with a relatively low level of complexity compared to SVM for the same feature combination. This model also exhibits a significant improvement in accuracy compared to other feature combinations for the K-NN model. The next combination of K-NN for the "Energy and Homogeneity" feature has a low level of complexity, and very high accuracy, reaching 99.98 %.

## 4. CONCLUSION

Based on the results of research related to the Gray Level Co-occurrence Matrix (GLCM) technique in feature extraction for object recognition, there are several important points that we can conclude. First, the GLCM method is proven to be effective in producing features that, when combined, can provide a high level of accuracy but with low complexity. This is especially crucial in the context of real-time applications such as security surveillance, autonomous car navigation, and so on. Furthermore, the K-Nearest Neighbors (K-NN) classification model shows a more efficient performance in terms of complexity when compared to the Support Vector Machine (SVM) model. In particular, K-NN, when using a combination of Correlation, Energy, and Homogeneity features, manages to achieve 100% accuracy with low complexity. Meanwhile, when using a combination of Energy and Homogeneity features, K-NN achieves an almost perfect accuracy level, namely 99.9889%, with an equally low level of complexity. On the other hand, even though SVM achieves 100% accuracy in some feature combinations, the high or very high complexity of this model can be a bottleneck, especially in the context of real-time applications. Thus, based on accuracy and complexity considerations, the K-NN model with a combination of the Correlation, Energy, and Homogeneity features appears to be a more suitable choice for real-time applications that require high levels of accuracy and low complexity.

## BIOGRAPHY OF AUTHORS

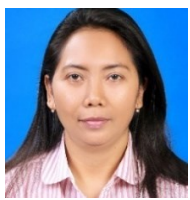

**Florentina Tatrin Kurniati** received her Master in Informatics Engineering from Atma Jaya Yogyakarta University (UAJY), Yogyakarta, Indonesia (2015), and is currently pursuing a doctoral program in computer science at Satya Wacana Christian University. Since 2008 he has been a lecturer and researcher at the faculty informatics and computers, Institut Teknologi dan Bisnis STIKOM Bali, Indonesia. He is interested in adaptive noise cancellation, pattern recognition, Object Identification, and digital forensics.

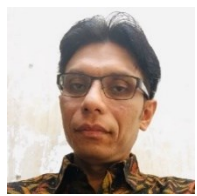

**Irwan Sembiring** completed his Bachelor's degree in 2001 at UPN "Veteran" Yogyakarta Indonesia. His Master degree is completed in 2004 from Gadjah Mada University Yogyakarta Indonesia, and Doctorate degree completed in 2016 at Gadjah Mada University Indonesia. Research interest in Network Security, Information System and Digital Forensic. Now he is a lecturer at faculty of Information Technology Satya Wacana Christian University, Salatiga, Indonesia.

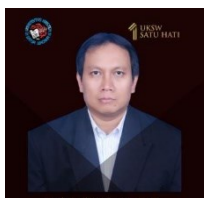

**Adi Setiawan** received his doctoral degree at the Vrije Universiteit Amsterdam, in the field of Statistics in 2007. Completed his master's degree at the Vrije Universiteit Amsterdam, in the field of Mathematics, and completed his bachelor's degree at Universitas Gadjah Mada, in the field of mathematics in 1991.

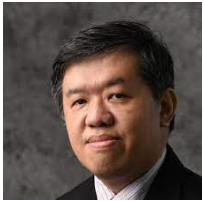

**Iwan Setyawan** received his Bachelor's and Master's degrees in 1996 and 1999, respectively, from the Department of Electrical Engineering, Bandung Institute of Technology, Indonesia. He received his Ph.D. degree from the Department of Electrical Engineering, Delft University of Technology, The Netherlands. He is currently an Assistant Professor at the Department of Electrical and Computer Engineering, Satya Wacana Christian University, Indonesia. His current research interest includes digital image and video compression and watermarking.

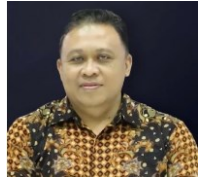

**Roy Rudolf Huizen** graduated with Doctor of Computer Science (2018) from Universitas Gadjah Mada (UGM) Yogyakarta, Indonesia. Lecturer and researcher at the Department of Magister Information System at the Institut Teknologi dan Bisnis STIKOM Bali, with research interests in the fields of Object Identification, Signal Processing, Cyber Security Forensics and Artificial Intelligence.